%% file: main.tex
\documentclass[conference]{IEEEtran}
\IEEEoverridecommandlockouts
\usepackage{cite}
\usepackage{amsmath,amssymb,amsfonts}
\usepackage{algorithmic}
\usepackage{graphicx}
\usepackage{textcomp}
\usepackage{xcolor}
\usepackage{hyperref}
\usepackage{pifont}
\usepackage{booktabs} 
\def\BibTeX{{\rm B\kern-.05em{\sc i\kern-.025em b}\kern-.08em
    T\kern-.1667em\lower.7ex\hbox{E}\kern-.125emX}}

\usepackage{subfiles}

\begin{document}

\subfile{our_article}
\subfile{sup}

\end{document}

%% file: our_article.tex
\title{GAMED-Snake: Gradient-aware Adaptive
Momentum Evolution Deep Snake Model for Multi-organ Segmentation}

\author{\textbf{Ruicheng Zhang}$^{1\dag}$, \textbf{Haowei Guo}$^{1\dag}$, \textbf{Zeyu Zhang}$^2$, \textbf{Puxin Yan}$^1$, \textbf{Shen Zhao}$^{1*}$\\
$^1$Sun Yat-sen University $^2$The Australian National University

\thanks{$^{\dag}$These authors contributed equally to this work.

$^{*}$Corresponding author: z-s-06@163.com.}}
\maketitle
\begin{abstract}
Multi-organ segmentation is a critical yet challenging task due to complex anatomical backgrounds, blurred boundaries, and diverse morphologies. This study introduces the Gradient-aware Adaptive Momentum Evolution Deep Snake (GAMED-Snake) model, which establishes a novel paradigm for contour-based segmentation by integrating gradient-based learning with adaptive momentum evolution mechanisms. The GAMED-Snake model incorporates three major innovations: First, the Distance Energy Map Prior (DEMP) generates a pixel-level force field that effectively attracts contour points towards the true boundaries, even in scenarios with complex backgrounds and blurred edges. Second, the Differential Convolution Inception Module (DCIM) precisely extracts comprehensive energy gradients, significantly enhancing segmentation accuracy. Third, the Adaptive Momentum Evolution Mechanism (AMEM) employs cross-attention to establish dynamic features across different iterations of evolution, enabling precise boundary alignment for diverse morphologies. Experimental results on four challenging multi-organ segmentation datasets demonstrate that GAMED-Snake improves the mDice metric by approximately 2\% compared to state-of-the-art methods.
Code will be available at \url{https://github.com/SYSUzrc/GAMED-Snake}.

\end{abstract}


\vspace{-1em}
\section{Introduction}
\label{sec:intro}
\vspace{-0.5em}
Multi-organ segmentation, which predicts the boundaries of all tissues of interest within an image, is of significant clinical value. Many organs are anatomically interconnected and functionally interdependent. Hence, their contours and morphology are often considered simultaneously for the diagnosis and treatment of certain diseases. For example, in radiation therapy (RT) for cancer, accurately delineating organs at risk (OARs) is crucial for minimizing the adverse effects. Typically, an RT session requires the segmentation of dozens of OARs, making manual segmentation labor-intensive and time-consuming. In contrast, automatic multi-organ segmentation can significantly reduce the required effort and time while enhancing the consistency, accuracy, and reliability of the results.

\begin{figure}[htbp]
    \centering
    \vspace{-0.15cm}
    \includegraphics[width=0.5\textwidth]{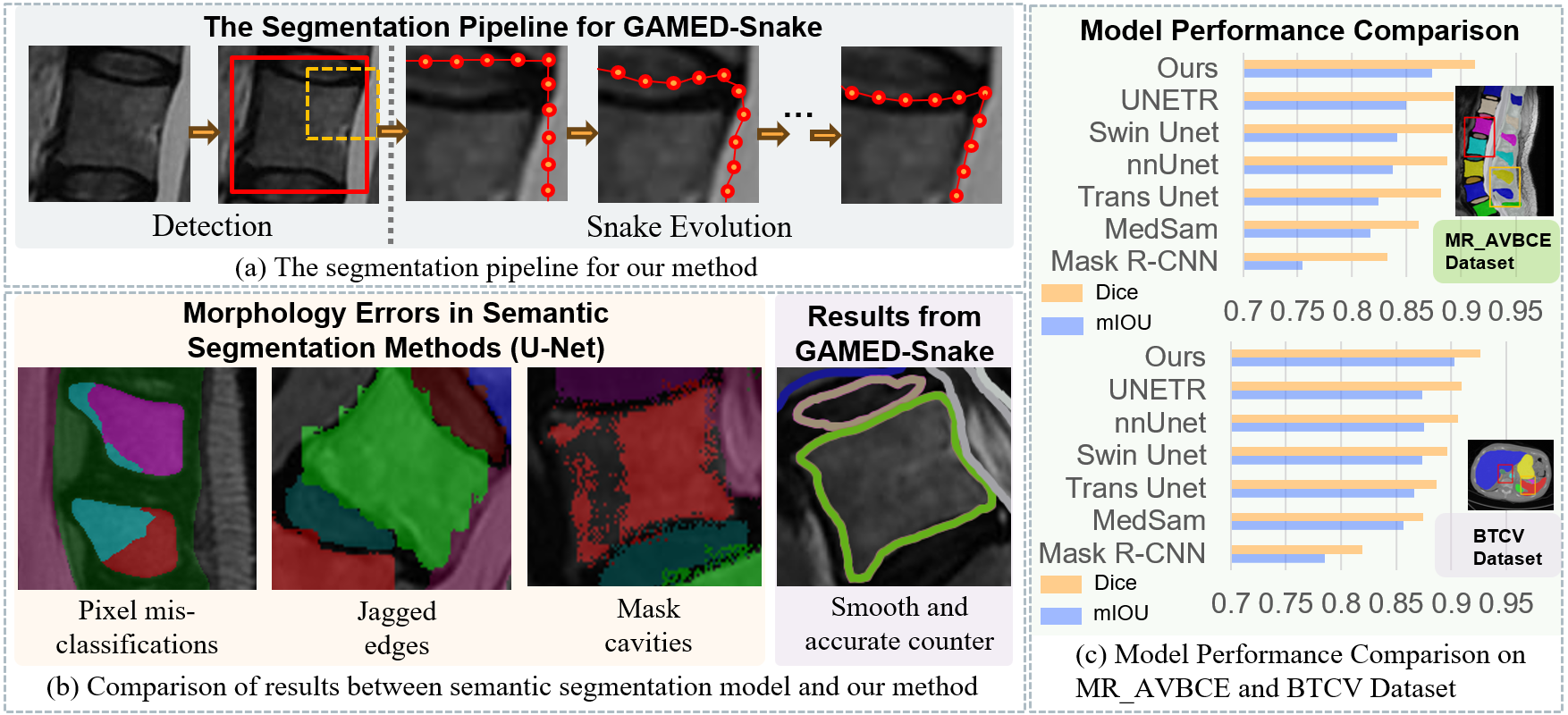}
    \setlength{\abovecaptionskip}{-0.45cm}   

\setlength{\belowcaptionskip}{-1.8cm}   
    \caption{(a) The workflow of GAMED-Snake consists of two stages: initialization of detection boxes and contour evolution. Taking the detection boxes as the initial contours, snake evolution process iteratively deforms them to match organ boundaries. (b) Semantic segmentation models based on pixel classification often struggle with complex multi-organ segmentation scenes, resulting in errors as illustrated in Fig. \ref{fig:image1}(b). In contrast, snake algorithms inherently avoid these issues, producing smooth and precise contours. (c) Improvement of GAMED-Snake over the SOTA approaches on MR\_AVBCE \cite{Zhao2023Attractive} and BTCV \cite{landman2015miccai} datasets.}
    \label{fig:image1}
    \vspace{-0.55cm}
\end{figure}

However, multi-organ segmentation remains a challenging task due to its complex nature \cite{zhang2024segreg}. First, complex backgrounds with numerous interfering structures make it more difficult to accurately identify and segment target organs. Second, the boundaries between adjacent organs are often blurred, and their tight anatomical arrangement further complicates precise contour delineation. Moreover, the wide diversity in the shapes and sizes of different organs poses significant challenges for a single model to generalize effectively. These challenges could hinder the efficacy of existing semantic segmentation methods \cite{tan2024segstitch}, most of which treat segmentation as a pixel-wise classification task \cite{wu2023bhsd,zhang2023thinthick}. These approaches fail to explicitly consider the global structure of the target organs at the object level, leading to a lack of holistic understanding. As a result, the segmentation outcomes are often inconsistent, exhibiting pixel misclassifications, jagged contours and mask cavities, as illustrated in Fig. \ref{fig:image1}.

\begin{figure*}[tbp]
    \centering
    \vspace{-0.11cm}
    \includegraphics[width=0.99\textwidth]{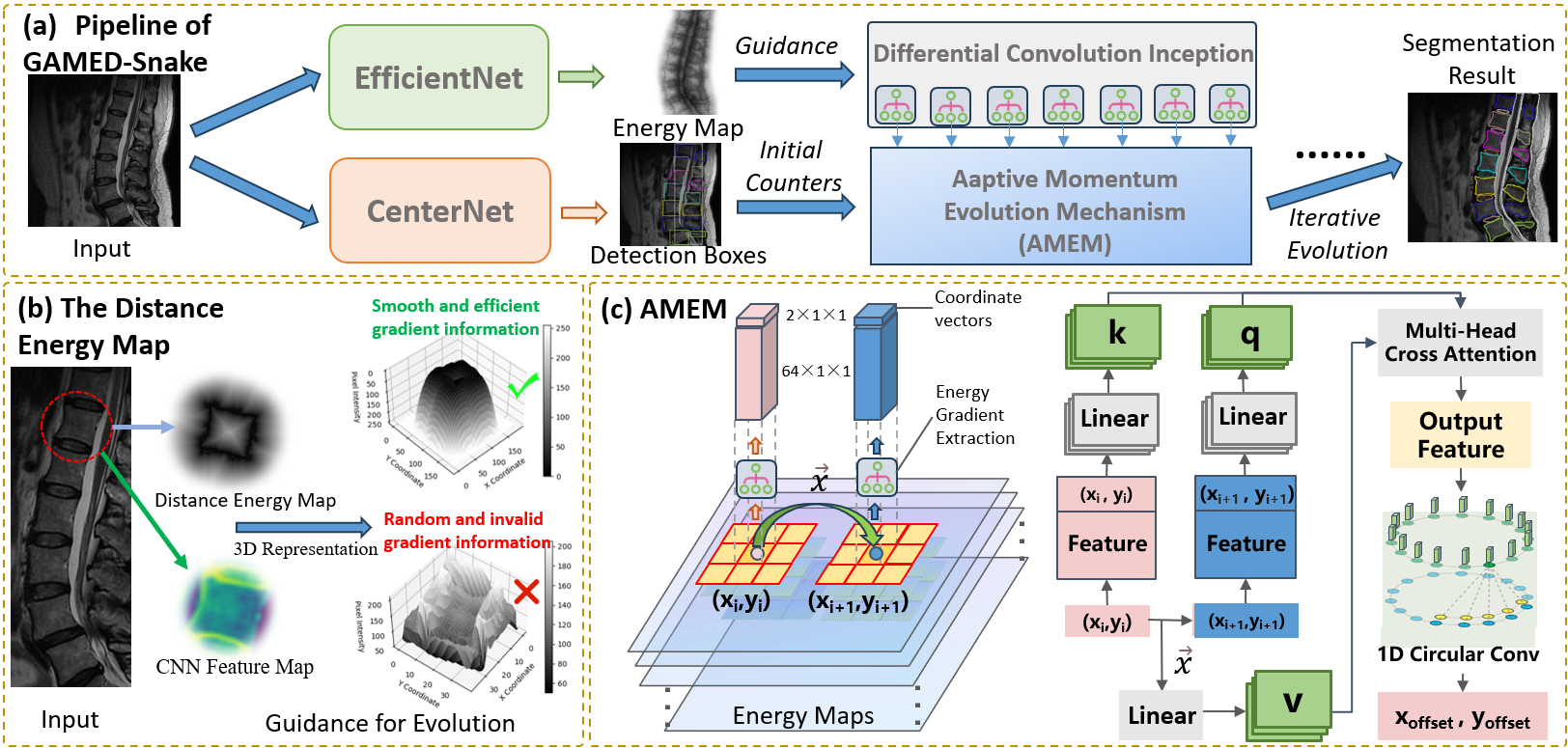}
    \setlength{\abovecaptionskip}{-0.05cm}   

\setlength{\belowcaptionskip}{-1.5cm}   
    \caption{(a) \textbf{The pipeline of GAMED-Snake}: GAMED-Snake first generates initial contours and then deforms them to align with the target boundaries under the guidance of energy maps. (b) \textbf{The principles underlying the Distance Energy Map}: This map encodes the distance distribution to guide contour evolution effectively. (c) \textbf{The structure of the Adaptive Momentum Evolution Mechanism (AMEM)}: AMEM adaptively integrates current and historical state information, establishing dynamic features across different iterations of evolution.}
    \label{fig:long}
    \vspace{-0.60cm}
\end{figure*}

Snake algorithm \cite{snake}, particularly when integrated with deep learning (i.e., deep snake), presents a promising solution to these challenges. Unlike conventional semantic segmentation algorithms \cite{tan2024segkan}, which predict pixel-level semantic maps \cite{ge2024esa}, the deep snake model generates initial object-level contours and refines them through vertex-wise offsets. This two-stage pipeline of detection followed by segmentation allows the model to focus on specific anatomical structures, mitigating the interference from complex backgrounds. Meanwhile, this object-level contour inherently accounts for the structural relationships among predicted regions, demonstrating robustness across diverse organ morphologies. Furthermore, the natural constraints between adjacent points ensure that the snake algorithm effectively produces smooth boundaries, even in cases of ambiguous edges, thereby avoiding the jagged and unrealistic contours common in pixel-based methods.

Nevertheless, evolving contour points effectively to accurately fit object boundaries is challenging, with previous methods achieving limited success, particularly in medical imaging. Most existing approaches \cite{snake,deep_snake,Xie_2020,Lazarow_2022} conceptualize the contour as a graph and employ graph convolutional networks to model snake evolution as a topological problem. While these frameworks offer a structured representation, they typically overlook the dynamic state-space transformation inherent in contour evolution, thereby limiting their effectiveness. Furthermore, the absence of prior anatomical knowledge may hinder these methods from accurately identifying the true boundaries in ambiguous medical images, resulting in suboptimal contours that are either  insufficiently fitted or excessively smoothed.

In this work, we propose the \textbf{G}radient-aware \textbf{A}daptive \textbf{M}omentum \textbf{E}volution \textbf{D}eep Snake (\textbf{GAMED-Snake}) model, introducing a novel paradigm for multi-organ segmentation. Our model leverages an innovative gradient-aware evolution strategy to guide the evolution process and incorporates an adaptive momentum attention mechanism. This mechanism enhances the ability of contour points to accurately locate target boundaries by dynamically perceiving evolutionary states.

GAMED-Snake utilizes the \textbf{D}istance \textbf{E}nergy \textbf{M}ap \textbf{P}rior (DEMP) to guide contour evolution, which encodes pixel-level distance information by intensifying pixel values as they approach the target contours. This design generates a strong force field across the image, effectively attracting contour points towards the target boundaries. Additionally, we design a novel \textbf{D}ifferential \textbf{C}onvolution \textbf{I}nception  \textbf{M}odule (DCIM) to effectively extract energy gradient information from the energy map, offering precise guidance on direction and step size for contour point evolution. Additionally, we propose an \textbf{A}daptive \textbf{M}omentum \textbf{E}volution \textbf{M}echanism (AMEM) to bolster the contour points' ability to search for boundaries in organs with varied morphologies. This mechanism adaptively integrates current and historical state information through cross-attention, establishing dynamic features across different iterations of evolution. Validation on four challenging multi-organ segmentation datasets demonstrates the superior performance of GAMED-Snake and its potential for clinical applications.

The contributions of this work are summarized as follows:

\begin{itemize}
 \item We propose a Gradient-aware Adaptive Momentum Evolution Deep Snake (GAMED-Snake) model for multi-organ segmentation. This model not only serves as a robust complement to semantic segmentation methods, but also offers novel insights into deep snake algorithms. 

\item GAMED-Snake employs a novel gradient-aware evolution strategy, leveraging the distance energy map as a strong prior to guide snake evolution. Combined with the differential convolution inception module for efficient energy gradient extraction, this strategy enhances robustness against complex backgrounds and ambiguous boundaries in multi-organ segmentation.

\item GAMED-Snake introduces an adaptive momentum evolution mechanism, utilizing an innovative cross-attention strategy to capture dynamic features between consecutive iterations. This enhances the ability of contour points to search for and align with target boundaries.

\end{itemize}

\vspace{-0.5em}
\section{RELATED WORKS}
\vspace{-0.25em}

\subsection{Multi-organ Segmentation}
\vspace{-0.31em}
Multi-organ segmentation is an essential and challenging task, attracting considerable research attention. Fang et al. \cite{Fang_Yan_2020} propose a multi-scale deep neural network incorporating pyramid convolution for multi-organ segmentation in CT images. Boutillon et al. \cite{Boutillon_2021} develop a segmentation model that utilizes shared convolutional kernels and domain-specific normalization for MRI images of three musculoskeletal joints. Shen et al. \cite{Shen__2023} introduce a spatial attention block that improves abdominal CT segmentation by learning spatial attention maps to highlight organs of interest. Zhao et al. \cite{Zhao_2022} combine a CNN-Transformer architecture \cite{cai2024msdet} with a progressive sampling module, achieving high performance in multi-organ segmentation for both CT and MRI images.

Despite these advances in segmentation accuracy, the inherent limitations of pixel-to-pixel prediction render these methods vulnerable to the challenges posed by complex backgrounds and ambiguous boundaries in multi-organ segmentation scenarios. Additionally, the lack of strong anatomical priors and the failure to explicitly account for the structural relationships among predicted outputs often result in unreasonable morphological errors, such as mask cavities, fragmented or jagged boundaries, and erroneous pixel classifications.

\vspace{-0.1cm}
\subsection{Deep Snake Algorithm}
Deep snake algorithms, which extend traditional active contour models \cite{snake} (ACMs) by incorporating deep learning techniques, demonstrate significant potential in multi-organ segmentation. By focusing on contour evolution rather than pixel-wise classification, these models are capable of generating smooth and realistic boundaries, even in scenarios with blurred edges or complex backgrounds. This makes deep snake methods a strong complement to semantic segmentation approaches, which also  motivates our work. Xie et al. \cite{Xie_2020} reformulate instance segmentation into instance center classification and dense distance regression tasks by modeling instance masks in polar coordinates. Peng et al. \cite{deep_snake} propose a two-stage deep snake pipeline that utilizes a novel circular convolution for efficient feature learning to enhance snake evolution. Lazarow et al. \cite{Lazarow_2022} introduce a point-based transformer with mask supervision via a differentiable rasterizer. However, these approaches typically treat snake evolution as a purely topological problem, overlooking its dynamic nature. Additionally, the absence of robust anatomical prior guidance may constrain their segmentation performance, particularly in the challenging multi-organ segmentation scenarios. 

Despite the potential of deep snake models to effectively parameterize object boundaries, their application remains underexplored, particularly within the medical imaging domain.

\vspace{-0.4em}
\section{METHOD}
Inspired by \cite{deep_snake}, GAMED-Snake performs segmentation by iteratively deforming an initial contour to align with the target organ boundary. Specifically, the method takes an initial contour as input and predicts vertex-wise offsets directed towards the target boundary, as depicted in Fig. \ref{fig:image1}. We propose an innovative contour evolution strategy that leverages the Differential Convolution Inception Module (DCIM) to effectively extract gradient information from the distance energy map. This gradient information offers precise guidance for determining the direction and step sizes for contour point offsets. In addition, the Adaptive Momentum Evolution Mechanism (AMEM) establishes dynamic features across successive evolution iterations, enhancing the ability of contour points to accurately search for the target boundary.

\vspace{-0.1cm}
\subsection{Distance Energy Map Prior}
Traditional snake algorithms rely on low-level image features such as grayscale gradients and predefined ACM parameters to guide contour evolution \cite{Shen__2023}. However, this weak guidance proves insufficient for handling the complexities of multi-organ medical images \cite{Zhao2023Attractive}, which frequently feature intricate backgrounds, blurred boundaries, and diverse contour morphologies.

To overcome these limitations, our GAMED-Snake incorporates a novel Distance Energy Map Prior (DEMP) within the deep snake framework, providing precise guidance for contour evolution. As a high-level feature map, the DEMP effectively encodes the distance of each pixel to the target contour boundary, offering a concise yet robust prior representation. For a given pixel $P(x, y)$, its energy value $E_P$ is defined based on the distance to its nearest boundary point $C(x, y)$:

\vspace{-0.3cm}
\begin{small}
\begin{equation}
    E_P = \max \left\{ 0, 255 - 32\ln \left( 1 + \| P(x, y) - C(x, y) \|_2 \right) \right\}.
\end{equation}
\end{small}
\vspace{-0.5cm}

The design generates a force field distributed across the entire image,  attracting contour points precisely to the target boundaries to achieve precise alignment.

\begin{figure}[tbp]
    \centering
    \vspace{-0.15cm}
    \includegraphics[width=0.5\textwidth]{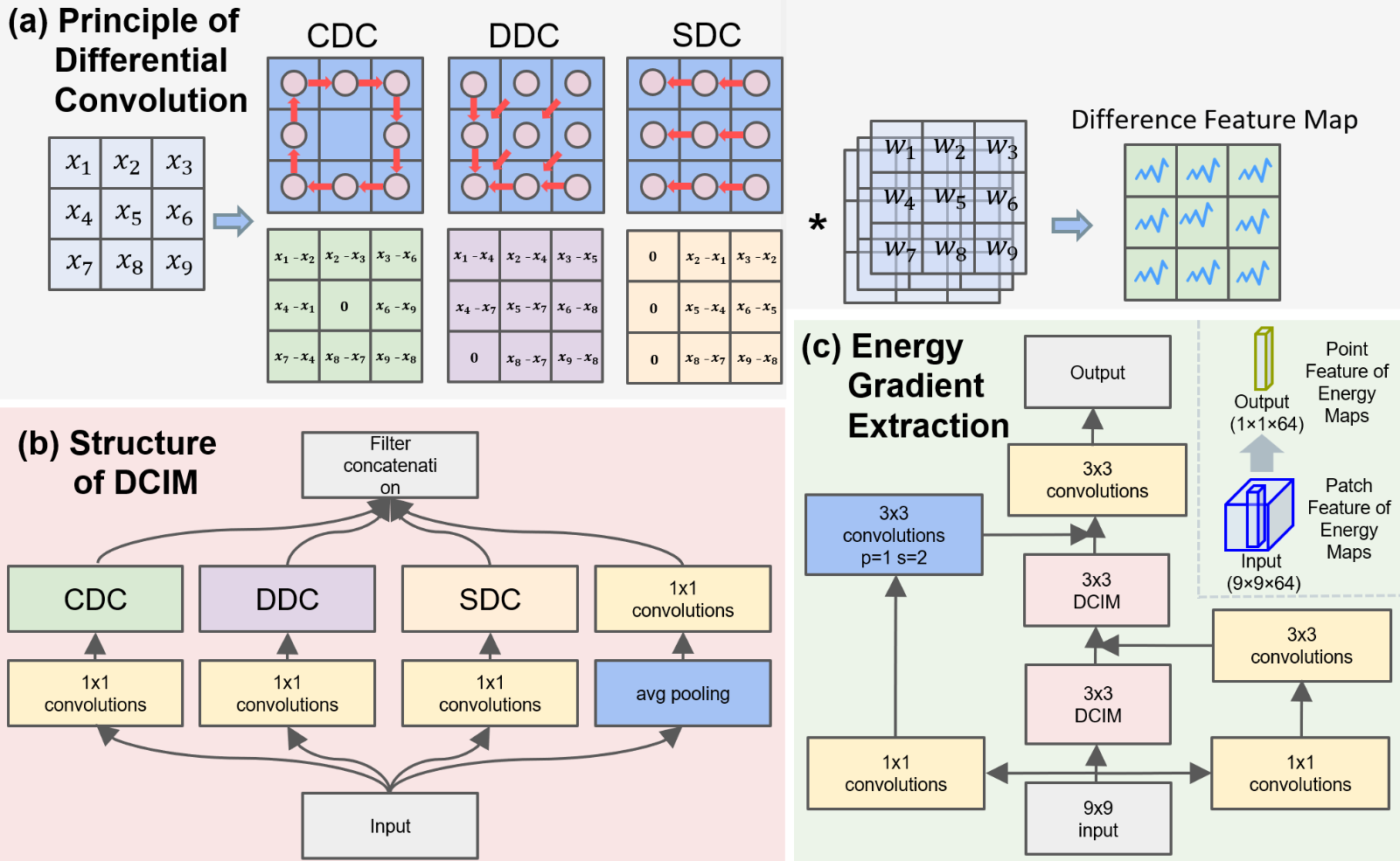}
    \setlength{\abovecaptionskip}{-0.45cm}   

\setlength{\belowcaptionskip}{-1.8cm}   
    \caption{(a) The working principle of differential convolution. (b) The structure of the Differential Convolution Inception Module (DCIM). (c) The process of energy gradient extraction. We aggregate feature information near contour point locations with DCIM to guide the snake evolution process.}
    \label{fig:dc}
    \vspace{-0.60cm}
\end{figure}

\vspace{-0.1cm}
\subsection{Differential Convolution Inception Module}
\vspace{-0.1cm}
In distance energy maps, gradient information can encode both the distance and direction of a given pixel relative to the target boundary. This information offers  effective guidance for the step size and direction of contour point evolution. Previous studies \cite{PDC} have shown that integrating traditional edge operators with convolutional neural networks significantly enhances gradient detection capabilities. Inspired by this, we design a Differential Convolution Inception Module (DCIM) to adaptively extract various types of energy gradient information.

The differential convolution (DC) process is similar to standard convolution (SC). During the convolution operation, instead of using the original pixel intensities, DC replaces them with pixel differences within the local feature map patch covered by the convolution kernels. This modification allows the network to focus on gradient changes instead of absolute intensities, thus enhancing its sensitivity to boundary features:

\vspace{-0.2cm}
\begin{small}
\begin{equation}
\begin{aligned}
& \text{SC}:\ y = f(x,\theta ) = {\mathop \sum \nolimits }_{i = 1}^{k \times k} {w_i} \cdot {x_i}, \\ & \text{DC}:\ y = f(\nabla x,\theta ) = {\mathop \sum \nolimits }_{\left( {{x_i},x_i^\prime } \right) \in \mathcal{P}} {w_i} \cdot \left( {{x_i} - x_i^\prime } \right), 
\end{aligned} 
\end{equation}
\vspace{-0.3cm}
\end{small}

 \noindent where $x_i$ and $x_i'$ represent the input pixels, and $w_i$ denote the weights in the $k \times k$ convolution kernel. 
The set $P=\{(x_1, x_1'), (x_2, x_2'), \ldots, (x_m, x_m')\}$ contains pixel pairs sampled from the current local patch, with $m \leq k \times k$.

To comprehensively capture gradient information, DCIM comprises four distinct branches: Stepped Differential Convolution (SDC), Diagonal Differential Convolution (DDC), Circular Differential Convolution (CDC), and average pooling. Each form of DC is used to capture different gradient information, providing comprehensive gradient awareness for contour points located at various spatial positions. The principles underlying each DC type are straightforward. For example, in CDC, pixel differences are calculated within a $3 \times 3$ patch along both diagonal and radial directions. These pixel differences are then element-wise multiplied by the kernel weights and convolved, followed by summation to produce the output feature map values (See Fig. \ref{fig:dc}).


\vspace{-0.1cm}
\subsection{Adaptive Momentum Evolution Mechanism}
\vspace{-0.1cm}
Regressing contour-point offsets in a single step is challenging, especially for vertices far away from the organ. Inspired by \cite{deep_snake,Xie_2020,Lazarow_2022}, we  handle contour evolution in an iterative optimization fashion. Previous methods typically treat the evolution process as a topological problem, largely neglecting its dynamic properties. However, in the context of state-space transition problems, the temporal dependencies between states are of paramount importance. To address this oversight, we design an Adaptive Momentum Evolution Mechanism (AMEM), which establishes dynamic features across adjacent evolution iterations, thereby effectively enhancing the contour points' ability to accurately locate target boundaries.

In the evolution process, vertex position offsets are predicted based on contour-point feature information. The input feature $\mathbf{f}_{\text{i}}$ for a vertex $x_i$ is a concatenation of learning-based features and the vertex coordinate: $[F(x_i); x_i]$, where $F$ denotes the feature maps. AMEM extracts feature vectors of the contour points at both current ($\mathbf{f}_{\text{c}}$) and historical ($\mathbf{f}_{\text{h}}$) positions, which are then fused using a cross-attention mechanism with historical evolution vectors $\mathbf{x}$:

\begin{small}
\begin{equation}
\begin{aligned}
\mathbf{q} &= \mathbf{f}_{\text{c}}^T \mathbf{W}_q, \quad \mathbf{k} = \mathbf{f}_{\text{h}}^T \mathbf{W}_k, \quad \mathbf{v} = \mathbf{x}^T \mathbf{W}_v, \\
\mathbf{A} &= \text{softmax}\left(\frac{\mathbf{q} \mathbf{k}^T}{\sqrt{C/h}}\right), \quad \text{CA}(\mathbf{f},\mathbf{x}) = \mathbf{A} \mathbf{v},
\end{aligned}
\end{equation}
\end{small}

 \noindent where \(\mathbf{W}_q, \mathbf{W}_k, \mathbf{W}_v \in \mathbb{R}^{C \times (C/h)}\) are learnable parameters, \(C\) and \(h\) denote the embedding dimension and the number of heads, respectively. 

The outputs are processed through multi-layer circular 1D convolutions \cite{deep_snake} to produce the final contour-point offset vectors. AMEM adaptively compresses information from both historical and current states, using the displacement vector from the previous step as “\textbf{\textit{momentum}}” to guide the current evolution step, significantly improving the ability of contour points to locate target boundaries. Moreover, the circular 1D convolution integrates features from neighboring points, effectively enlarging the receptive field of contour evolution.

\vspace{-0.1cm}
\subsection{Implementation details}

\textbf{The energy map generation network}  \ The distance energy map generation network is built upon the EfficientNetV2 \cite{EfficientNetV2} backbone, followed by deconvolution layers for outputting predictions. EfficientNet optimizes network architecture through neural architecture search, significantly boosting performance with a reduction in the number of parameters.

\textbf{Detector} \ We adopt CenterNet \cite{centernet} as the detector for our GAMED-Snake, which outputs class-specific boxes as the initial contours. CenterNet reformulates the detection task as a keypoint detection problem and achieves an impressive trade-off between speed and accuracy.

\textbf{Contour evolution} \ We uniformly sample $N$ points from both the ground truth boundary and the initial contour and pair them by minimizing the distance between corresponding poins. GAMED-Snake takes the initial contour as input and outputs $N$ offsets that point from each vertex to the target boundary point. We set $N$ to 128 in all experiments, which is sufficient to cover most organ shapes. The number of evolutionary iterations is set to 3.

\textbf{Training strategy} \ We initially pretrain the energy map generation network to ensure accurate distance energy map predictions. This is followed by the joint optimization of both the detection and the snake evolution processes. 

In the pretraining phase of the distance energy map, we utilize the Charbonnier loss, given by:

\vspace{-0.11cm}
\begin{small}
\begin{equation}
    \mathcal{L}_E = \sqrt{\left\| f_E(P(x, y)) - E_P^{GT} \right\|^2 + \epsilon^2}, \quad \epsilon = 10^{-3},
\end{equation}
\end{small}
\vspace{-0.51cm}

 \noindent where $E_P^{GT}$ denotes the distance energy value of the ground truth, and $f_E(\cdot)$ represents the energy map generation network.

Subsequently, we employ the smooth $L_1$ loss to train the detection and segmentation processes. The loss function for the prediction of the detection box is defined as:

\vspace{-0.2cm}
\begin{small}
\begin{equation} L_{ex}=\frac{1}{4}\sum_{i=1}^{4}\ell_{1}(\tilde{\mathbf{x}}_{i}^{ex}-\mathbf{x}_{i}^{ex}),  
\end{equation}
\end{small}
\vspace{-0.31cm}

 \noindent where $\tilde{\mathbf{x}}_{i}^{ex}$ and $\mathbf{x}_{i}^{ex}$ represent the predicted and actual vertices of the detection box, respectively. The loss function for iterative contour deformation is defined as:

\vspace{-0.21cm}
\begin{small}
\begin{equation} 
L_{iter}=\frac{1}{N}\sum_{i=1}^{N}\ell_{1}(\tilde{\mathrm{x}}_{i}-\mathrm{x}_{i}^{gt}), \quad N=128,
\end{equation}
\end{small}
\vspace{-0.31cm}

 \noindent where $\tilde{\mathrm{x}}_{i}$ is the deformed contour point and $\mathrm{x}_{i}^{gt}$ is the ground-truth boundary point. For the detection part, we adopt the same loss function as the original detection model.

\section{EXPERIMENTS}
\subsection{Experimental Settings}
\subsubsection{Dataset Introduction}
For our experiments, we utilize four multi-organ datasets, including the private multi-organ spinal dataset MR\_AVBCE \cite{Zhao2023Attractive}  and three public datasets: the spinal dataset VerSe \cite{sekuboyina2021verse}, the abdominal multi-organ segmentation dataset BTCV \cite{landman2015miccai}, and RAOS \cite{ma2023raos}.

The MR\_AVBCE dataset is a multi-organ segmented spinal dataset with 600 slices that includes vertebrae, intervertebral discs, the spinal cord, and other attachments. The VerSe dataset is a large-scale, multi-device, and multi-center CT spine segmentation dataset, comprising data from the VerSe19 and VerSe20 Challenges at MICCAI 2019 and 2020. The BTCV dataset is an abdominal organ segmentation benchmark that involves 13 different organs, such as the spleen, kidneys, gallbladder, liver, stomach, and others. The RAOS dataset features a broader range of organs, including 19 distinct organs. We employ a slicing technique on 3D sequences to generate a dataset comprised of 2D slices. For further details, please refer to the supplementary materials.

\subsubsection{Evaluation Metrics}
We evaluate the model's segmentation performance using two metrics: mean Intersection over Union (mIoU) and mean Dice score (mDice). Specifically, IoU defined as \(\text{IoU}(X, X^*) = \frac{\vert X \cap X^* \vert}{\vert X \cup X^* \vert}\), where $X^*$ denotes the ground truth set, $X$ denotes the predicted segmentation set and $\vert X \vert$ denotes the number of pixels in $X$. Dice evaluates the similarity between $X^*$ and $X$ based on their overlap, given by \( \text{Dice}(X, X^*) = \frac{2 \times \vert X \cap X^* \vert}{\vert X \vert + \vert X^* \vert} \). 


\begin{table}[]
\vspace{-1em}
\centering
\setlength{\tabcolsep}{4pt}  
\renewcommand{\arraystretch}{1.0}  
\caption{Experiments on MR\_AVBCE \cite{Zhao2023Attractive}, VerSe \cite{sekuboyina2021verse}, BTCV \cite{landman2015miccai}, and RAOS \cite{ma2023raos} datasets. Optimal and suboptimal metric values are \textbf{bolded} and \underline{underlined}, respectively. }
\label{tab:t1} 
\vspace{-1.0em}
\scalebox{0.82}{  
\begin{tabular}{c|cc|cc|cc|cc}  
Datasets   & \multicolumn{2}{c|}{MR\_AVBCE} & \multicolumn{2}{c|}{VerSe} & \multicolumn{2}{c|}{BTCV} & \multicolumn{2}{c}{RAOS}  \\ 
Metrics & mIoU   & mDice   & mIoU   & mDice   & mIoU   & mDice  & mIoU   & mDice  \\ \midrule
nnUnet \cite{isensee2021nnunet} & 0.8366  & 0.8871 & \underline{0.8524}  & \underline{0.8879} & \underline{0.8746}  & 0.9058  & \underline{0.8789}  & \underline{0.8985} \\
UNETR \cite{hatamizadeh2022unetr}   &  \underline{0.8495} & \underline{0.8926} &   0.8377 & 0.8621  &   0.8737 & \underline{0.9095} &   0.8689 & 0.8846 \\
Trans Unet \cite{chen2021transunet}    & 0.8235 & 0.8811  & 0.8376 & 0.8563  & 0.8658 & 0.8865  & 0.8476 & 0.8786 \\
Swin Unet \cite{cao2021swinunet}   & 0.8412 & 0.8921 & 0.8489 & 0.8715& 0.8703 & 0.8968 & 0.8552 & 0.8847 \\
MedSam \cite{ma2024segment}       & 0.8162 & 0.8612  & 0.8273 & 0.8673  & 0.8565 & 0.8742  & 0.8595 & 0.8839 \\
Mask R-CNN \cite{he2017maskrcnn}       & 0.7542 & 0.8324 & 0.7032 & 0.7549 & 0.7846 & 0.8191 & 0.8067 & 0.8445\\
\midrule 
Ours       &    \textbf{0.8726} & \textbf{0.9123}  & \textbf{0.8835}  & \textbf{0.9011}  & \textbf{0.9027}  & \textbf{0.9264}  & \textbf{0.8945} & \textbf{0.9236}     \\ 
\midrule 
\end{tabular}
}
\end{table}

\begin{figure}[htbp]
    \centering
    \vspace{-0.35cm}
    \includegraphics[width=0.47\textwidth]{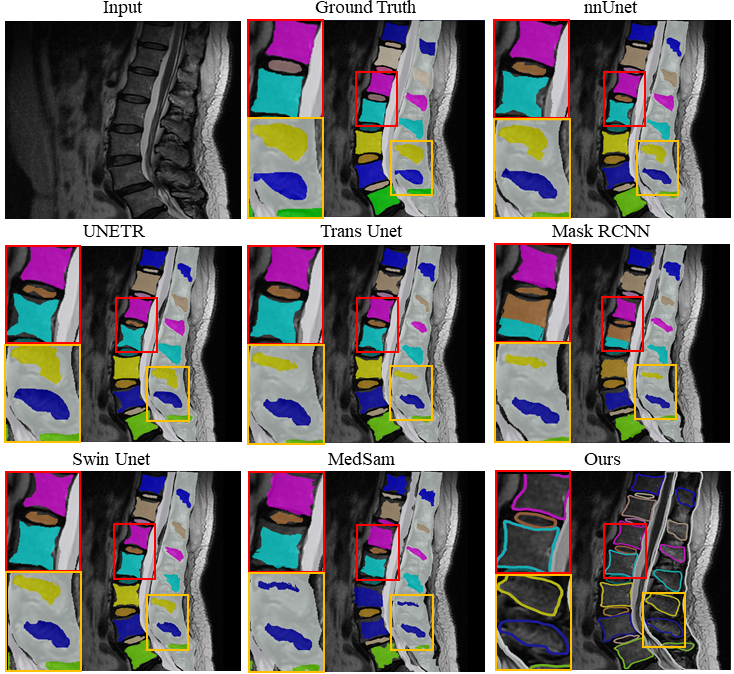}
    \setlength{\abovecaptionskip}{-0.15cm}   

\setlength{\belowcaptionskip}{-0.8cm}   
    \caption{Qualitative comparison of results on MR\_AVBCE datasets.}
    \label{fig:MR_AVBCE}
    \vspace{-0.20cm}
\end{figure}

\vspace{-0.2cm}
\subsection{Comparing Experiments}
\vspace{-0.1cm}
We conducted a comprehensive evaluation of GAMED-Snake against state-of-the-art (SOTA) and mainstream medical image segmentation models, including nnU-Net \cite{isensee2021nnunet}, UNETR \cite{hatamizadeh2022unetr}, TransUNet \cite{chen2021transunet}, SwinUnet \cite{cao2021swinunet}, MedSAM \cite{ma2024segment}, and Mask R-CNN \cite{he2017maskrcnn}.

\subsubsection{Quantitative Evaluation}
As shown in Table \ref{tab:t1}, GAMED-Snake consistently outperforms SOTA models across all four datasets. On MR\_AVBCE, GAMED-Snake surpasses the second-best models by 2.72\% in mIoU and 2.21\% in mDice. On the VerSe spinal dataset, the model demonstrates substantial improvements, with mIoU scores 3.65\% higher and mDice scores 1.49\% higher than the second-best methods.For abdominal datasets, GAMED-Snake achieves state-of-the-art performance. On BTCV, it achieves an average IOU of 0.9027, reflecting a 3.21\% improvement over nnU-Net, and an average Dice score of 0.9264, exceeding UNETR by 1.86\%. On RAOS, GAMED-Snake attains an average IOU of 0.8945, 1.67\% higher than nnU-Net, and an average Dice score of 0.9236, representing a 2.79\% improvement over nnU-Net.

\vspace{0.3cm}
\begin{figure}[hbp]
    \centering
    
    \includegraphics[width=0.5\textwidth]{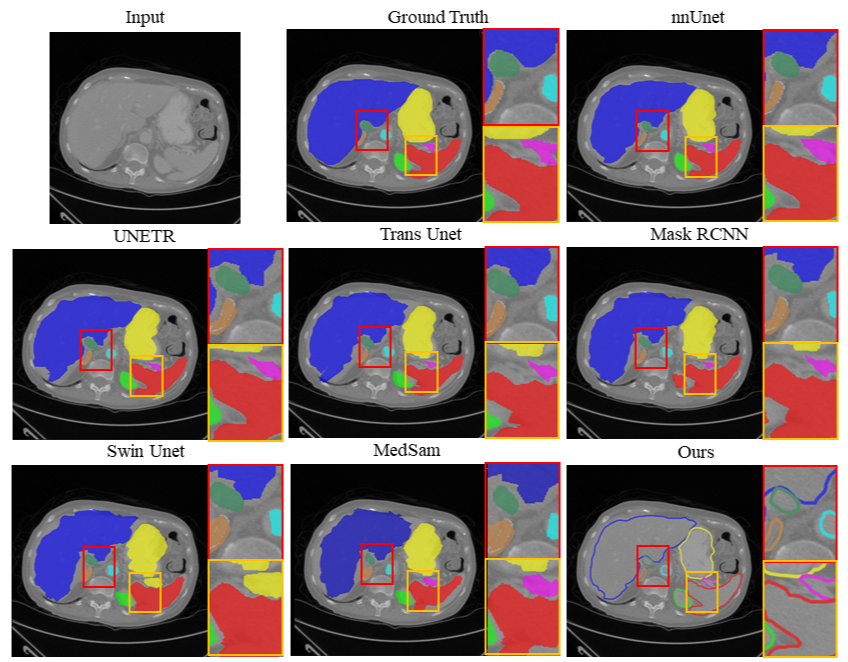}
    \setlength{\abovecaptionskip}{-0.55cm}   

\setlength{\belowcaptionskip}{-0.8cm}   
    \caption{Qualitative comparison of results on BTCV datasets.}
    \label{fig:BTCV}
    
\end{figure}

\vspace{-0.25cm}
\subsubsection{Qualitative Evaluation}

Qualitative comparisons are presented in Figs. \ref{fig:MR_AVBCE}, \ref{fig:BTCV} and \ref{fig:RAOS}. As depicted in Fig. \ref{fig:MR_AVBCE}, GAMED-Snake outperforms other methods in segmenting spinal multi-organ structures. Notably, when addressing adjacent vertebrae with highly similar appearances, pixel-wise semantic segmentation methods such as Mask RCNN \cite{he2017maskrcnn} and MedSAM \cite{ma2024segment} frequently exhibit inconsistent classification within the same tissue. In contrast, our method considers the holistic structural integrity of objects, effectively avoiding such errors. Additionally, the boundaries of vertebrae and the spinal cord, especially the spinous processes, were segmented more smoothly and accurately, whereas the segmentation results of other models showed discrepancies from the ground truth boundaries. 

Fig. \ref{fig:BTCV} and Fig. \ref{fig:RAOS} present the results of different methods for abdominal multi-organ segmentation. GAMED-Snake  mitigates issues such as jagged edges and mask cavities observed in other semantic segmentation models. Moreover, in scenarios involving closely arranged multi-organ structures, GAMED-Snake produces smoother and more natural boundary delineations, particularly for overlapping organ boundaries. In contrast, the segmentation results from other semantic segmentation models are often fragmented and irregular.

\vspace{-0.1cm}
\subsection{Ablation Study}
\vspace{-0.1cm}
We perform ablation studies on various architectural configurations of our model to investigate the contribution of each component on the MR\_AVBCE dataset. The detector generates object boxes, forming ellipses around them, which are then refined towards boundaries using Graph-ResNet. The results (Table \ref{tab:ablation}) demonstrate that both the DEMP\&DCIM and the AMEM significantly improve segmentation performance with their combination achieving the optimal results.

\begin{table}[h]
\vspace{-1.5em}
\setlength{\abovecaptionskip}{0cm}  
\setlength{\belowcaptionskip}{-0.2cm}
\centering
\small 
\caption{Ablation Experiments}
\label{tab:ablation}
\setlength{\abovecaptionskip}{-0.85cm}   
\resizebox{0.45\textwidth}{!}{  
\renewcommand{\arraystretch}{0.8}  
\begin{tabular}{*{4}{c}}  
    \toprule
      DEMP\&DCIM  &  AMEM &  mIoU &  mDice  \\
    \midrule
{\normalsize \ding{55}} & {\normalsize \ding{55}}  & {\normalsize 0.7986} & {\normalsize 0.8224}  \\
    {\normalsize \ding{55}}  & {\normalsize \ding{51}}  & {\normalsize$0.8525(\textcolor{red}{6.75\%}\uparrow)$} & {\normalsize $0.8894(\textcolor{red}{8.15\%}\uparrow)$}  \\
    {\normalsize \ding{51}}  & {\normalsize \ding{55}}  & {\normalsize $0.8467 (\textcolor{red}{6.02\%}\uparrow)$} & {\normalsize $0.8785(\textcolor{red}{6.82\%}\uparrow)$} \\
    {\normalsize \ding{51}}  & {\normalsize \ding{51}}  & {\normalsize $0.8726 (\textcolor{red}{9.27\%}\uparrow)$} & {\normalsize $0.9123(\textcolor{red}{10.93\%}\uparrow)$}  \\
    \bottomrule
\end{tabular}
}
\vspace{-1em}
\end{table}

\begin{figure}[htbp]
    \centering
    \vspace{-0.35cm}
    \includegraphics[width=0.45\textwidth]{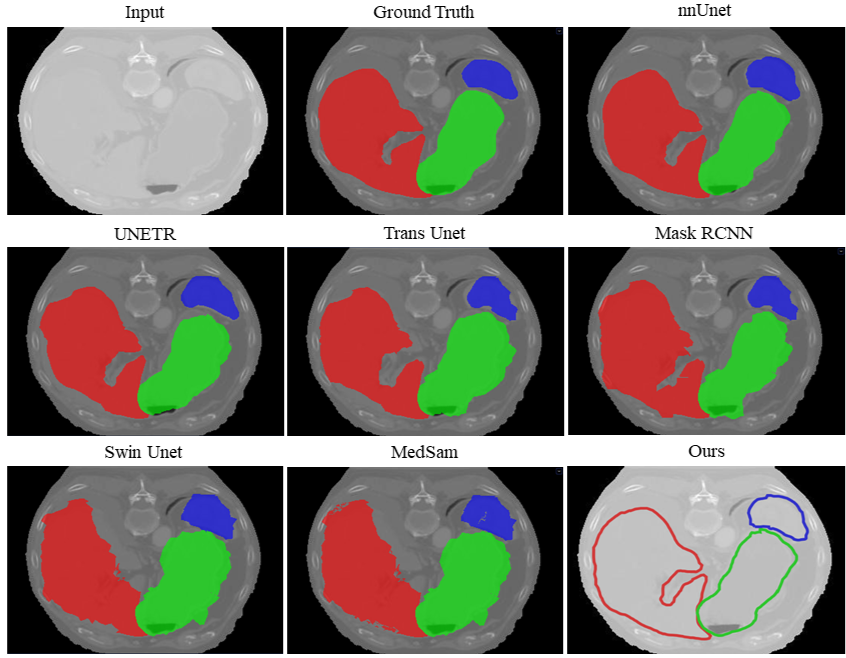}
    \setlength{\abovecaptionskip}{-0.15cm}   

\setlength{\belowcaptionskip}{-0.8cm}   
    \caption{Qualitative comparison of results on RAOS datasets.}
    \label{fig:RAOS}
    \vspace{-0.25cm}
\end{figure}

\section{CONCLUSION}
We introduce the GAMED-Snake model, a novel approach for multi-organ segmentation that integrates gradient-aware learning and adaptive momentum evolution into a unified contour-based framework. GAMED-Snake not only advances the design of snake algorithms but also functions as a robust complement to semantic segmentation methods. Utilizing the Distance Energy Map Prior and the Differential Convolution Inception Module, GAMED-Snake provides precise guidance for contour evolution, overcoming challenges posed by complex backgrounds and blurred boundaries. The Adaptive Momentum Evolution Mechanism establishes dynamic awareness across different evolution iterations, improving contour points' accuracy in locating boundaries of organs with diverse morphologies. We found that integrating boundary features with contour coordinates provides valuable guidance for precise segmentation and robust anatomical priors enhance the model's adaptability to complex features in medical images. These offer meaningful insights for future research.

 \vspace{-0.12cm}


%% file: sup.tex
\setcounter{section}{0}
\setcounter{table}{0}
\title{Supplementary}
\author{}

\maketitle
\section{introduction to the private dataset MR\_AVBCE}
The MR\_AVBCE dataset \cite{Zhao2023Attractive} consists of 600 MRI images, which are collected from three different medical institutions: Affiliated Hangzhou First People's Hospital, Qilu Hospital of Shandong University, and Saint Joseph Health Care Center in London. These images are acquired using different imaging techniques, including T1-weighted and T2-weighted imaging 
 \cite{bushong2013mri}, with an approximately equal number of each. Additionally, the MRI scans are conducted using equipment from different manufacturers, including GE, Siemens, and United Imaging. Consequently, the MR\_AVBCE dataset is heterogeneous, which increases the complexity and diversity of the data, more accurately reflecting the complexities encountered in clinical practice and enhancing the generalization capability of the trained models.

The MR\_AVBCE dataset contains 4,601 vertebrae, and a portion of the vertebrae in the dataset exhibit pathological deformations (approximately 820 vertebrae), primarily caused by tumors and degenerative diseases. Additionally, around 20 vertebrae suffer from imaging artifacts, and approximately 270 vertebrae have blurred vertebral body edges, resulting in low-quality images. The challenge lies in whether the model can effectively learn from these minority samples. MR\_AVBCE also captures the intricate image characteristics typically encountered in clinical practice, such as the considerable variability in vertebral sizes, intensity distributions, and pathological morphological deformations among different patients.

\section{implementation details}
\subsection{Data Processing}
CT and MR modalities generate 3D images, while other modalities, such as X-ray and ultrasound, produce 2D images \cite{ma2024segment}. To achieve broad applicability across various modalities, we designed GAMED-Snake as a general-purpose 2D segmentation model. This design allows it to process both 2D and 3D images by converting 3D volumes into a series of 2D slices. 

In the experiments, we extract 2D slices from the 3D volumes for analysis. Specifically, the VerSe dataset  \cite{sekuboyina2021verse} contains approximately 60 slices per CT scan, the BTCV dataset \cite{landman2015miccai} contains between 100 and 200 slices per CT scan, and the RAOS dataset \cite{ma2023raos} contains around 200 slices per CT scan. For the VerSe dataset, we use sagittal slices \cite{smith2014ct} for both training and evaluation, while for the abdominal datasets BTCV and RAOS, axial slices \cite{smith2014ct} are used. All slices are uniformly cropped to a resolution of 512 × 512 pixels.

\subsection{Experimental Setup}
The model is implemented in Python 3.7 with PyTorch 1.9.0, and all experiments are conducted on an NVIDIA RTX 3090 GPU. The Adam optimizer is employed for optimization. To enhance data diversity, data augmentation techniques such as flipping and rotation are applied during training. The batch size is set to 24, with an initial learning rate of 0.0001, which decays every 50 epochs at a rate of gamma = 0.5. The model is trained for a total of 200 epochs.

\begin{figure}[htbp]
    \centering
    \includegraphics[width=0.5\textwidth]{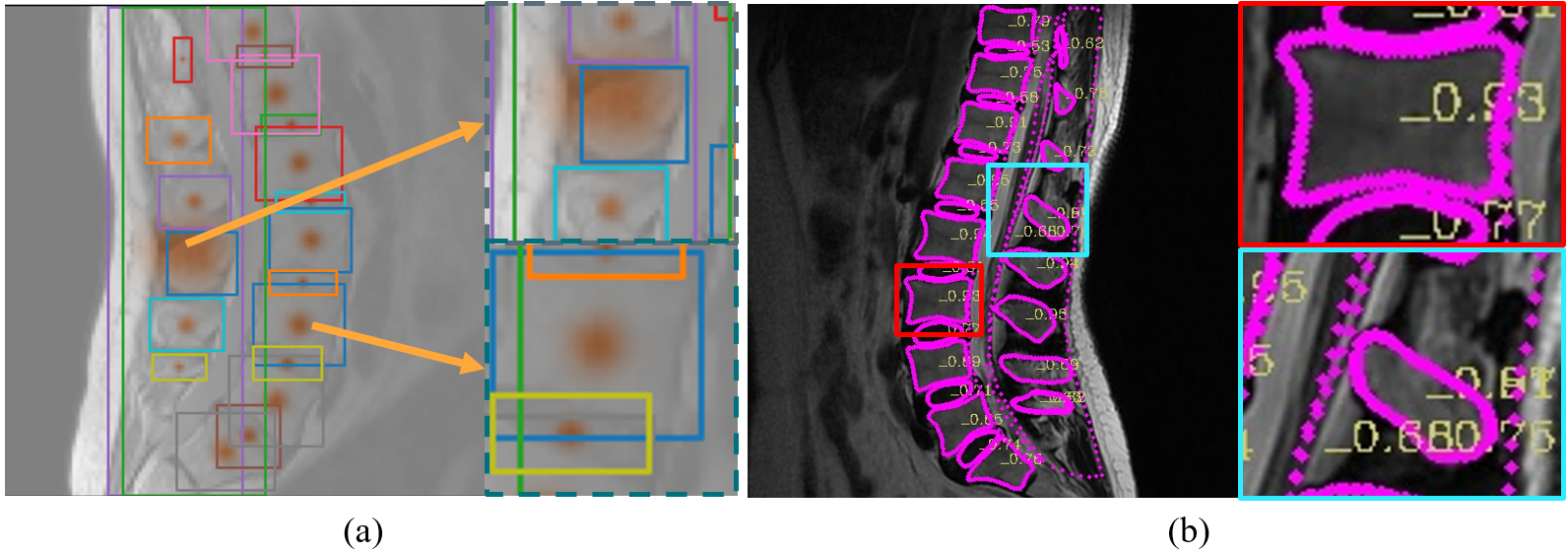}
    
    \caption{
(a) Output results from the detector (CenterNet \cite{centernet}), which includes the predicted heatmap of organ center points and detection boxes with associated class labels. The detection boxes serve as the initial contours for subsequent contour evolution. (b) Visualization of contour points. The results demonstrate that 128 points are sufficient to cover various target boundaries in multi-organ segmentation scenarios.}
    \label{fig:img1}
\end{figure}

\subsection{Number of Sampling Points} 

In our model, the number of contour points is set to 128, which is sufficient to cover the majority of organ and tissue boundaries in multi-organ segmentation scenarios. As illustrated in Figure 1, a contour comprising 128 points can fully capture smaller targets, such as the vertebra, without any loss of segmentation accuracy. Moreover, it forms smooth and precise contours around elongated structures, such as the spinal cord. Experimental results further confirm that 128 points represent the minimum number required to maintain optimal performance. As shown in Table \ref{tab:t1}, reducing the number of points to 64 results in a significant decline in model performance, while increasing the number to 192 or 256 does not lead to noticeable performance gains.  All experiments were conducted on the MR\_AVBCE dataset.

\begin{table}[htbp]

\centering
\setlength{\tabcolsep}{4pt}  
\renewcommand{\arraystretch}{1.5}  
\caption{Comparison of results for different numbers of sampling points.}
\label{tab:t1} 
\vspace{-1.0em}
\scalebox{0.92}{  
\begin{tabular}{c c c c c c}  
\hline
Number of Sampling Points & \multicolumn{1}{c }{64} & \multicolumn{1}{c }{96} & \multicolumn{1}{c }{128} & \multicolumn{1}{c }{192} & \multicolumn{1}{c}{256}   \\ 
\hline
mIoU                 & 0.7834   &0.8351  & 0.8726 & 0.8727  & 0.8715                                        \\ 
mDice                & 0.8363      &0.8968                & 0.9123        &0.9125              & 0.9121                                      \\ 
\hline
\end{tabular}
}
\end{table}

\subsection{Number of Evolution Iterations}

In the experiments, we analyze the impact of varying evolution iteration counts on model performance by retraining and testing the model using different configurations. As shown in Table \ref{tab:t2}, an iteration count of 3 achieves the highest mIoU and mDice scores (0.9123), indicating an optimal balance between accuracy and computational efficiency. Increasing the number of iterations to 4 or 5 slightly reduces performance metrics, likely due to the increased training complexity and potential overfitting. All experiments were conducted on the MR\_AVBCE dataset.

\begin{table}[htbp]

\centering
\setlength{\tabcolsep}{4pt}  
\renewcommand{\arraystretch}{1.5}  
\caption{Comparison of results for different numbers of evolution iterations.}
\label{tab:t2} 
\vspace{-1.0em}
\scalebox{0.92}{  
\begin{tabular}{c c c c c c}  
\hline
Number of Iterations & \multicolumn{1}{c }{Iter. 1} & \multicolumn{1}{c }{Iter. 2} & \multicolumn{1}{c }{Iter. 3} & \multicolumn{1}{c }{Iter. 4} & \multicolumn{1}{c}{Iter. 5}  \\ 
\hline
mIoU                 & 0.8226                      & 0.8678                      & 0.8726                      & 0.8714  & 0.8708                    \\ 
mDice                & 0.8563                      & 0.9086                      & 0.9123                      & 0.9013  & 0.9011                    \\ 
\hline
\end{tabular}
}
\end{table}

\section{Detector Introduction}
The detector in GAMED-Snake utilizes CenterNet \cite{centernet}, an anchor-free object detection algorithm that predicts the center point location of the object along with its associated attributes, such as width, height, and class, to achieve efficient object detection and classification. Unlike traditional anchor-based detection methods, CenterNet locates objects by generating a center point heatmap and then regresses the object's bounding box dimensions, simplifying and enhancing the accuracy of the detection process.

In GAMED-Snake, CenterNet adopts DLA-34 \cite{DLA} as its backbone network. The output layer comprises three branches: heatmap, offset, and size, with corresponding output dimensions of \((W/R, H/R, C)\), \((W/R, H/R, 2)\), and \((W/R, H/R, 2)\), respectively, where \(R\) denotes the stride (set to 4 in this study) and \(C\) represents the number of organ classes. CenterNet identifies the center points of objects within the image, allowing the model to focus on target organs while mitigating interference caused by unclear boundaries and complex backgrounds. Additionally, the bounding box dimensions and aspect ratios provide rough morphological cues about the segmented objects, enabling the model to better adapt to organs with significant variations in shape and size.

As an anchor-free object detection framework, CenterNet supports efficient end-to-end training. Its straightforward center point prediction and regression mechanism not only accurately locates targets in medical images but also seamlessly integrates with snake evolution. This integration provides improved guidance for delineating target regions, thereby enhancing segmentation accuracy.

\bibliographystyle{IEEEbib}
\bibliography{main}